\documentclass{article} % For LaTeX2e
\usepackage{nips15submit_e,times}
\usepackage{hyperref}
\usepackage{url}
\usepackage{algorithm,algorithmic}
\usepackage{amsmath}
\usepackage{amssymb}
\usepackage{graphicx}
\usepackage{caption}
\usepackage{subcaption}
\usepackage{amsthm}

\title{A Scale Mixture Perspective of Multiplicative Noise in Neural Networks}

\author{
Eric T. Nalisnick\\
Department of Computer Science\\
University of California, Irvine\\
\texttt{enalisni@uci.edu}
\AND
Anima Anandkumar \\
Dept. of Electrical Eng. and Comp. Sci. \\
University of California, Irvine \\
\texttt{a.anandkumar@uci.edu} \\
\And
Padhraic Smyth \\
Department of Computer Science\\
University of California, Irvine\\
\texttt{smyth@ics.uci.edu} %\\
}

\nipsfinalcopy % Uncomment for camera-ready version

\begin{document}
\maketitle
\begin{abstract}
Corrupting the input and hidden layers of deep neural networks (DNNs) with multiplicative noise, often drawn from the Bernoulli distribution (or `dropout'), provides regularization that has significantly contributed to deep learning's success.  However, understanding how multiplicative corruptions prevent overfitting has  been difficult due to the complexity of a DNN's functional form.  In this paper, we show that when a Gaussian prior is placed on a DNN's weights, applying multiplicative noise induces a Gaussian scale mixture, which can be reparameterized to circumvent the problematic likelihood function.  Analysis can then proceed by using a type-II maximum likelihood procedure to derive a closed-form expression revealing how regularization evolves as a function of the network's weights.  Results show that multiplicative noise forces weights to become either sparse or invariant to rescaling.  We find our analysis has implications for model compression as it naturally reveals a weight pruning rule that starkly contrasts with the commonly used signal-to-noise ratio (SNR). While the SNR prunes weights with large variances, seeing them as noisy, our approach recognizes their robustness and retains them.  We empirically demonstrate our approach has a strong advantage over the SNR heuristic and is competitive to retraining with soft targets produced from a teacher model.
\end{abstract}

\section{Introduction}
Training deep neural networks (DNNs) under multiplicative noise, by introducing a random variable into the inner product between a hidden layer and a weight matrix, has led to  significant improvements in predictive accuracy.  Typically the noise is drawn from a Bernoulli distribution, which is equivalent to randomly dropping neurons from the network during training, and hence the practice has been termed \textit{dropout} \cite{hinton2012improving, srivastava2014dropout}. Recent work \cite{srivastava2014dropout, tomczak2013prediction} suggests equivalent, if not better, performance   using  Beta or Gaussian distributions for the multiplicative noise.  Thus, in this paper we consider  multiplicative noise regularization broadly, not limiting our focus  just to the Bernoulli distribution.

Despite its empirical success, regularization by way of multiplicative noise is not well understood theoretically, especially for DNNs.  The multiplicative noise term eludes analysis as a result of being buried within the DNN's composition of non-linear functions. In this paper, by adopting a Bayesian perspective, we show that we can develop closed-form analytical expressions that describe the effect of training with multiplicative noise in DNNs and other models.  When a zero-mean Gaussian prior is placed on the weights of the DNN, the multiplicative noise variable induces a \textit{Gaussian scale mixture} (GSM), i.e. the variance of the  Gaussian prior  becomes a random variable whose distribution is determined by the  multiplicative noise model.
Conveniently, GSMs can be represented hierarchically with the scale mixing variable---in this case the multiplicative noise---becoming a hyperprior. This allows us to     circumvent the problematic coupling of the noise and likelihood through reparameterization,  making them conditionally independent.  Once in this form a type-II maximum likelihood procedure yields closed-form updates for the multiplicative noise term and hence makes the regularization mechanism explicit.

While the GSM reparameterization and learning procedure are not novel in their own right, employing them to understand multiplicative noise in neural networks is new.  Moreover, the analysis is not restricted by the network's depth or activation functions, as previous attempts at understanding dropout have been.  We show that regularization via multiplicative noise has a dual nature, forcing weights to become either sparse or invariant to rescaling.  This result is consistent with, but also expands upon, previously-derived adaptive regularization penalties for linear and logistic regression \cite{wager2013dropout}.  
 
As for its practical implications, our analysis suggests a new criterion for principled model compression.  The closed-form regularization penalty isolated herein naturally suggests a new weight pruning strategy. Interestingly, our new rule is in stark disagreement with the commonly used \textit{signal-to-noise ratio} (SNR) \cite{graves2011practical, blundell2015weight}.  The SNR is quick to prune weights with large variances, deeming them noisy, but our approach finds large variances to be an essential characteristic of robust, well-fit weights.   
Experimental results on well-known predictive modeling tasks show that our weight pruning mechanism is not only superior to the SNR criterion by a wide margin,  but also competitive to retraining with soft-targets produced by the full network \cite{hinton2014distilling, ba2014deep}.  In each experiment our method was able to prune at least 20\% more of the model's parameters than SNR before seeing a vertical asymptote in test error.  Furthermore, in two of these experiments, the performance of models pruned with our method reduced or matched the error rate of the retrained networks until reaching 50\% reduction.
 
\section{Dropout Training and Previous Work}\label{Dropout_sec}
Below we establish notation for training under multiplicative noise (MN) and review some relevant previous work on dropout.  In general, matrices are denoted by bold, upper-case variables, vectors by bold, lower-case, and scalars by both upper and lower-case.    Consider a neural network with $L$ total layers ($L-2$ of them hidden).  Forward propagation consists of recursively computing \begin{equation}\label{layer_compute}
\mathbf{h}_{l} = f_{l}(\mathbf{h}_{l-1}\mathbf{W}_{l})
\end{equation} where $\mathbf{h}_{l}$ is the $d_{l}$-dimensional vector of hidden units located at layer $l$, $\mathbf{h}_{l-1}$ is the $d_{l-1}$-dimensional vector of hidden units located at the previous layer $l-1$, $f_{l}$ is some (usually non-linear) element-wise activation function associated with layer $l$, and $\mathbf{W}_{l}$ is the $d_{l-1}\times d_{l}$-dimensional weight matrix.  If $l-1=1$, then $\mathbf{h}_{1-1}=\mathbf{x}_{i}$, a vector of input features corresponding to the $i$th training example out of $N$, and if $l=L$, then $\mathbf{h}_{L}=\hat y_{i}$, the class prediction for the $i$th example.  For notational simplicity, we'll assume the bias term is absorbed into the weight matrix and a constant is appended to $\mathbf{h}_{l-1}$.    Training a neural network consists of minimizing the negative log likelihood: $\mathcal{L} = \sum_{i=1}^{N}-\log \hat y_{i} = -\log p(\mathbf{y} | \mathbf{X}, \mathbf{W})$ where $p(\mathbf{y} | \mathbf{X}, \mathbf{W})$ is a conditional distribution parameterized by the neural network.  $\mathbf{W}$ is learned through the backpropagation algorithm.

\subsection{Training with Multiplicative Noise}
Training with multiplicative noise (MN) is a regularization procedure implemented through slightly modifying Equation (\ref{layer_compute}).  It causes the intermediate representation $\mathbf{h}_{l-1}$ to become stochastically corrupted by introducing random variables to the inner product $\mathbf{h}_{l-1}\mathbf{W}_{l}$.  Rewriting Equation (\ref{layer_compute}) with MN, we have \begin{equation}\label{dropout_compute}
\mathbf{h}_{l} = f_{l}(\mathbf{h}_{l-1}\boldsymbol{\Lambda}_{l}\mathbf{W}_{l})
\end{equation} where $\boldsymbol{\Lambda}_{l}$ is a diagonal $d_{l-1}\times d_{l-1}$-dimensional matrix of random variables $\lambda_{j,j}$ drawn independently from some noise distribution $p(\lambda)$.  Dropout corresponds to a Bernoulli distribution on $\lambda$  \cite{hinton2012improving, srivastava2014dropout}.

Training proceeds by sampling a new $\boldsymbol{\Lambda}_{l}$ matrix for every forward propagation through the network.  Backpropagation is done as usual using the corrupted values.  We can view the sampling as Monte Carlo integration over the noise distribution, and therefore, the MN loss function can be written as \begin{equation}\label{dropout_loss}
\mathcal{L}_{\text{MN}} = \mathbb{E}_{p(\lambda)}[-\log p(\mathbf{y} | \mathbf{X}, \mathbf{W},\boldsymbol{\Lambda})]
\end{equation} where the expectation is taken with respect to the noise distribution $p(\lambda)$.  At test time, the bias introduced by the noise is corrected; for instance, the weights would be multiplied by $(1-p)$ if we trained with Bernoulli($p$) noise.

\subsection{Closed-Form Regularization Penalties}\label{wager}
Direct analysis of Equation (\ref{dropout_loss}) for neural networks with non-linear activation functions is currently an open problem.  Nevertheless, analysis of dropout has received a significant amount of attention in the recent literature, and progress has been made by considering second order approximations \cite{wager2013dropout}, asymptotic assumptions \cite{wang2013fast}, linear networks \cite{baldi2013understanding, warde2013empirical}, generative models of the data \cite{wager2014altitude}, and convex proxy loss functions \cite{HelmboldL14}.

Since this paper is primarily concerned with interpreting MN regularization as a closed-form penalty, we
summarize below the results of \cite{wager2013dropout}, which had similar goals, in order to build on them later.  A closed-form regularization penalty can be derived exactly for linear regression and approximately for logistic regression.  For linear regression, training under MN is equivalent to training with the following penalized likelihood \cite{wager2013dropout, wang2013fast, baldi2013understanding}: \begin{equation}\begin{split}\label{clt_dropout} \mathcal{L}_{\text{MN LR}} &= \sum_{i=1}^{N}(y_{i}-\mathbf{x}_{i} \mathbf{w} )^{2} + \frac{1}{2}\mathrm{Var}[\lambda] \sum_{j=1}^{d}w_{j}^{2}\sum_{i=1}^{N}x_{i,j}^{2}. \end{split}\end{equation}  The second term can be viewed as data-driven $\ell_2$ regularization in that the weights are being penalized not by just their squared value but also by the sum of the squared features in the corresponding dimension.  Similarly, an approximate closed-form objective can be found for logistic regression via a 2nd-order Taylor expansion around the mean of the noise \cite{wager2013dropout}: \begin{equation}\begin{split}\label{taylor_dropout} \mathcal{L}_{\text{MN LogR}} &\approx -\sum_{i=1}^{N} y_{i}\log f(\mathbf{x}_{i}\mathbf{w}) + (1-y_{i})\log (1-f(\mathbf{x}_{i}\mathbf{w})) \\ &+ \frac{1}{2}\mathrm{Var}[\lambda] \sum_{j=1}^{d}w_{j}^{2}\sum_{i=1}^{N}f(\mathbf{x}_{i}\mathbf{w})(1-f(\mathbf{x}_{i}\mathbf{w}))x_{i,j}^{2}.
\end{split}\end{equation}  Again we find an $\ell_2$ penalty adjusted to the data and, in this case, the model's current predictions.  However, Helmbold and Long \cite{HelmboldL14} have suggested that this approximation can substantially underestimate the error.

\section{Multiplicative Noise as an Induced Gaussian Scale Mixture}
In this section below we go beyond prior work to show that analysis of multiplicative noise (MN) regularization can be made tractable by adopting a Bayesian perspective.  The key observation is that if we assume the weights to be Gaussian random variables, the product $\lambda w$, where $\lambda $ is the noise and $w$ is a weight, defines a \textit{Gaussian scale mixture} (GSM).  GSMs can be represented hierarchically with the scale mixing variable---in this case the noise $\lambda$---becoming a hyperprior.  The reparameterization works even for deep neural networks (DNNs) regardless of their size or activation functions.

\subsection{Gaussian Scale Mixtures}
First we define a Gaussian scale mixture.  A random variable $\theta$ is a Gaussian scale mixture (GSM) if and only if it can be expressed as the product of a Gaussian random variable--call it $u$--with zero mean and some variance $\sigma_{0}^{2}$ and an independent scalar random variable $z$ \cite{andrews1974scale,beale1959scale}:
\begin{equation}\label{GSM_def}
\theta \,{\buildrel d \over =}\, zu
\end{equation} where $\,{\buildrel d \over =}\,$ denotes equality in distribution.  While it may not be obvious from (\ref{GSM_def}) that $\theta$ is a \textit{scale} mixture, the result follows from the Gaussian's closure under linear transformations, resulting in the following marginal density of $\theta$: \begin{equation}\label{marginal}\begin{split}
p(\theta) &= \int p(\theta|z) p(z) dz \\
&= \int N(0,\sigma_{0}^{2}z^{2})p(z)dz
\end{split}\end{equation} where $p(z)$ is the mixing distribution.  Super-Gaussian distributions, such as the Student-t ($z^{2} \sim \text{Inverse Gamma}$), can be represented as GSMs, and this hierarchical formulation is often used when employing these distributions as robust priors \cite{steel2000bayesian}.

Now that we've defined GSMs, we demonstrate how MN can give rise to them.  Consider the addition of a Gaussian prior to the MN training objective given in Equation (\ref{dropout_loss}): \begin{equation*}
\mathcal{L}_{\text{GSM}} = \mathbb{E}_{p(\lambda)}[- \log (p(\mathbf{y} | \mathbf{X}, \mathbf{W},\boldsymbol{\Lambda})p(\mathbf{W}))]
\end{equation*} where, for a DNN, $p(\mathbf{W}) = \prod_{l=1}^{L-1}\prod_{j=1}^{d_{l-1}} \prod_{k=1}^{d_{l}} N(0,\sigma_{0}^{2})$, i.e., an independent Gaussian prior on each weight coefficient with some constant variance $\sigma_{0}^{2}$.  Next recall the inter-layer computation defined in Equation (\ref{dropout_compute}): $$\mathbf{h}_{l} = f_{l}(\mathbf{h}_{l-1}\boldsymbol{\Lambda}_{l}\mathbf{W}_{l}) = f_{l}(\mathbf{a}_{l}).$$  $\mathbf{a}_{l}$ is a $d_{l}$ dimensional vector whose $k$th element can be written in summation notation as $$a_{l,k} = \sum_{j=1}^{d_{l-1}} h_{l-1,j}\lambda_{l,j}w_{l,j,k}.$$  Notice that $w_{l,j,k} \sim N(0,\sigma_{0}^{2})$ and $\lambda_{l,j} \sim p(\lambda_{j})$; thereby making the product $\lambda_{l,j}w_{l,j,k}$ the definition of a GSM given in (\ref{GSM_def}).

The result follows just from application of the definition, but for a more intuitive explanation,  consider the case of a constant $c$ multiplied by a Gaussian random variable $w \sim N(0, \sigma_{0}^{2})$ as above.  The product $cw$ is distributed as $N(0,c^{2}\sigma_{0}^{2})$ due to the Gaussian's closure under scalar transformation.  The definition of a GSM (\ref{GSM_def}) says that the same result holds even if $c$ is a random variable---the only difference being the variance $c^{2}\sigma_{0}^{2}$ is now random itself.  See \cite{andrews1974scale} and \cite{beale1959scale} for rigorous treatments.

\subsection{The Hierarchical Parameterization for DNNs}\label{hier_sec}
\begin{figure}
\centering
\begin{subfigure}{.49\textwidth}
  \centering
  \includegraphics[width=.89\linewidth]{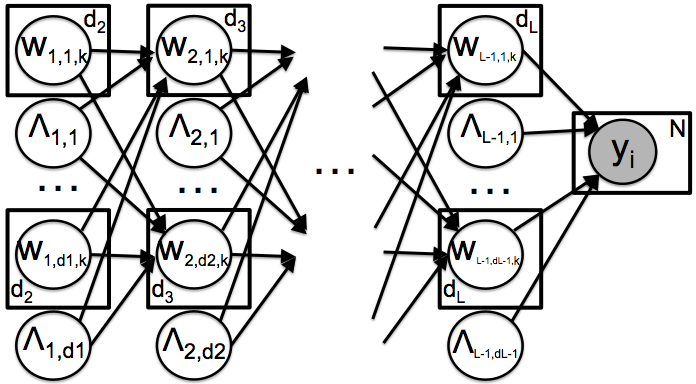}
  \caption{Unidentifiable (Multiplicative Noise)}
  \label{fig:sub0p}
\end{subfigure}
\begin{subfigure}{.49\textwidth}
  \centering
  \includegraphics[width=.89\linewidth]{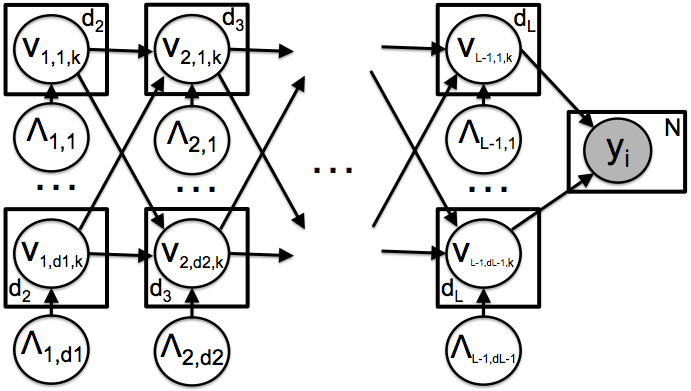}
  \caption{Identifiable (Hierarchical)}
  \label{fig:sub1p}
\end{subfigure}
\caption{Equivalent GSM parameterizations for a deep neural network.  It is distributionally equivalent to replace the product $\lambda w \sim N(0, \lambda^{2})$ with a new random variable $v \sim N(0, \lambda^{2})$.  The noise's ($\lambda$) influence is preserved, flowing through the weight, $v$, instead of directly into the next layer.}
\label{params}
\end{figure}

Here we introduce a key insight: the product between the weights of a DNN and the noise can be represented hierarchically, as given in Equation (\ref{marginal}), making the intractable likelihood conditionally independent of the noise.  Again, the reparameterization follows from the definition, and it can be seen graphically in Figure \ref{params}.  But to elaborate, it's equivalent (in distribution) to replacing the product  $\lambda_{l,j}w_{l,j,k} \sim N(0, \sigma_{0}^{2}\lambda_{l,j}^{2})$ with a new conditionally Gaussian random variable $v_{l,j,k} \sim N(0, \sigma_{0}^{2}\lambda_{l,j}^{2})$,   with $\lambda_{l,j}$ drawn from the noise distribution.  The random rescaling that $\lambda_{l,j}$ explicitly applied to $w_{l,j,k}$ is still present yet collapsed into the distribution from which $v_{l,j,k}$ is drawn\footnote{Just like it is equivalent, in the previous example using the constant $c$, to represent the distribution of $cw$ with a random variable $w^{*}\sim N(0,c^{2}\sigma_{0}^{2})$.}.  Because this interaction occurs entirely within the activation function, the complexities it introduces do not come into play.  The only dependence that needs to be accounted for when reparameterizing is the shared variance of all weights occupying the same row of $\mathbf{W}_{l}$ (due to the noise being sampled for each hidden unit).  This poses no serious complications   and is actually a desirable property, as we  discuss later.  From here forward, the product form of a GSM is referred to as the \textit{unidentifiable} parameterization---since only the product $\lambda_{l,j}w_{l,j,k}$ can be identified in the likelihood---and the hierarchical form the \textit{identifiable} parametrization.

\subsection{Dropout's Corresponding Prior}
We now turn to the case of $\lambda \sim$ Bernoulli($p$), the most widely used noise distribution.  Moving the Bernoulli random variable to the Gaussian random variable's scale reveals the classic prior for Bayesian variable selection, the \textit{Spike and Slab} \cite{mitchell1988bayesian,george1993variable}: \begin{equation}
p( v_{l,j,k} | \lambda_{l,j}) = \begin{cases}
    \delta_{0} & \text{if } \lambda_{l,j}=0\\
    N(0,\sigma_{0}^{2})              & \text{if } \lambda_{l,j}=1
\end{cases}
\end{equation} where $\delta_{0}$ is the delta function placed at zero.  Interestingly, the unidentifiable parameterization has been used previously for linear regression in the work of Kuo and Mallick \cite{kuo1998variable}.  They placed the Bernoulli indicators directly in the likelihood as follows, $$y_{i} = \sum_{j=1}^{p} \beta_{j}\gamma_{j}x_{i,j} + \epsilon_{i}, $$ where $\gamma_{j} \sim \text{Bernoulli}(p=0.5)$, essentially defining dropout for linear regression over a decade before it was proposed for neural networks.  However, Kuo and Mallick were interested in the marginal posterior inclusion probabilities $p(\gamma_{j}=1 | \mathbf{y})$ rather than predictive performance.

\section{Type-II ML for the Hierarchical Parameterization}\label{em_alg_sec}
Having established $p(\mathbf{y} | \mathbf{X}, \mathbf{W},\boldsymbol{\Lambda})p(\mathbf{W})$ can be written as $p(\mathbf{y} | \mathbf{X}, \mathbf{V})p(\mathbf{V}|\boldsymbol{\Lambda})$, we next wish to isolate the characteristics of the weights encouraged by multiplicative noise (MN) regularization.  Our aim is to write $\boldsymbol{\Lambda}$ as a function of $\mathbf{V}$ so we can explicitly see the interplay between the noise and parameters.  To do this, we  learn $\boldsymbol{\Lambda}$ from the data via a type-II maximum likelihood procedure (a form of empirical Bayes).  Note that this is hard to do in the unidentifiable parameterization due to explaining away \cite{murphy2012machine}.  The identifiable (hierarchical) parameterization, on the other hand, allows for an Expectation-Maximization\footnote{Actually, we  perform an equivalent minimization, instead of maximization, in the M-step to keep notation consistent with earlier equations.} (EM) formulation, as described in \cite{tipping2001sparse}.  The derivation of the EM updates is as follows:   \begin{equation}\label{em3}\begin{split}
\mathcal{L} &= - \log p(\boldsymbol{\Lambda} | \mathbf{y}, \mathbf{X}) \\ &\propto - \log [p(\mathbf{y} | \mathbf{X}, \boldsymbol{\Lambda})p(\boldsymbol{\Lambda})] \\ &\le  \int q(\mathbf{V}) -\log \frac{p(\mathbf{y} | \mathbf{X}, \mathbf{V})p(\mathbf{V}|\boldsymbol{\Lambda})p(\boldsymbol{\Lambda})}{q(\mathbf{V})} d\mathbf{V}.
\end{split}\end{equation}  We   make two simplifying assumptions to make working with the posterior manageable.  The first is, following \cite{tipping2001sparse}, we  choose $q(\mathbf{V})=p(\mathbf{V} | \mathbf{y}, \mathbf{X}, \boldsymbol{\hat \Lambda} )$, which corresponds to approximating the joint posterior with $p(\mathbf{V}, \boldsymbol{\Lambda} | \mathbf{y}, \mathbf{X}) \approx p(\mathbf{V} | \mathbf{y}, \mathbf{X}, \boldsymbol{\Lambda}) \delta_{\text{MAP}}(\boldsymbol{\Lambda})$.  The second assumption is that $p(\mathbf{V} | \mathbf{y}, \mathbf{X}, \boldsymbol{\Lambda})$ factorizes over its dimensions.

Hence, the E-Step is computing \begin{equation} Q_{t} = \mathbb{E}_{\mathbf{V} | \mathbf{y}, \mathbf{X}, \boldsymbol{\Lambda}}[-\log p(\mathbf{V}|\boldsymbol{\Lambda})p(\boldsymbol{\Lambda})],\end{equation} where the likelihood was dropped since it doesn't depend on $\boldsymbol{\Lambda}$, and the M-Step is \begin{equation}\begin{split}
\hat \lambda^{t+1}_{l,j} &=  \arg\min_{\lambda_{l,j}} Q_{t}.\end{split}\end{equation}  In our case, $p(\mathbf{V}| \mathbf{\Lambda})$ is a fully-factorized Gaussian so the gradient is \begin{equation}\label{m_grad}\begin{split}
\frac{\partial Q_{t}}{\partial \lambda_{l,j}} &=  \frac{\sum_{k=1}^{d_{l}}\mathbb{E}_{v | \mathbf{y}, \mathbf{X}, \boldsymbol{\Lambda}}[v_{l,j,k}^{2}]}{\lambda_{l,j}^{3}}-\frac{d_{l}}{\lambda_{j,k}} + \frac{\partial}{\partial \lambda_{l,j}}p(\lambda_{l,j}).
\end{split}\end{equation}  Unfortunately, the EM formulation cannot handle discrete noise distributions (and by extension, discrete mixtures) since we can't calculate $\frac{\partial Q_{t}}{\partial \lambda_{l,j}}$ if $\lambda_{l,j}$ is not a continuous random variable.  While this does not allow us to address Bernoulli noise (i.e. dropout) exactly, this is not a severe limitation for a few reasons. Firstly, as   discussed later,  the noise distribution encourages particular values for $\lambda_{l,j}$ but does not fundamentally change the nature of the regularization being applied to the DNN's weights. Secondly, empirical observations support that our conclusions apply to Bernoulli noise as well. Lastly, the Beta($\alpha$,$\beta$) with $\alpha=\beta<1$ can serve as a continuous proxy for the Bernoulli($0.5$).

\section{Analysis of the Regularization Mechanism}
Equation (\ref{m_grad}) provides an important window into the effect of multiplicative noise (MN) by revealing the properties of the weights that influence the regularization.  Below we analyze Equation (\ref{m_grad}) in detail, showing that multiplicative noise results in weights becoming either sparse or invariant to rescaling.  We start by setting (\ref{m_grad}) to zero, making the substitution $\mathbb{E}[v^{2}]=\mathbb{E}^{2}[v] + \text{Var}[v]$, and rearranging to solve for the variance term: \begin{equation}\label{type_2_soln}
\hat \lambda^{2}_{l,j} = \frac{1}{d_{l}}\sum_{k=1}^{d_{l}}\mathbb{E}_{v | y, \lambda}^{2}[v_{l,j,k}] + \frac{1}{d_{l}}\sum_{k=1}^{d_{l}}\text{Var}_{v | y, \lambda}[v_{l,j,k}] + \frac{\partial}{\partial \lambda_{l,j}}p(\lambda_{l,j}).
\end{equation}  The first term is the squared posterior mean, and the second is the posterior variance.  Both are averaged across weights emanating from the same unit due to the dependence discussed in Section \ref{hier_sec}.  The third term is the derivative of the noise distribution.  Moreover, notice that the $\frac{\partial}{\partial \lambda_{l,j}}p(\lambda_{l,j})$ term does not contain the DNN's parameters and therefore only serves as a prior expressing which values of $\lambda_{l,j}$ are preferred.  The regularization pertinent to the network's parameters is contained in the first two terms only.

In light of this observation, we  discard the noise distribution term for the time being and work with just the first two empirical Bayesian terms.  We can substitute them into the  variance of the Gaussian prior on $\mathbf{V}$ to see what regularization penalty MN is applying, in effect, to the weights: \begin{equation}\label{r_term_marg}\begin{split}
\mathcal{R}_{GSM}(\mathbf{V}) &= - \log p(\mathbf{V}| \boldsymbol{\hat \Lambda}) \\ &= \frac{1}{\sigma_{0}^{2}}\sum_{l=2}^{L}\sum_{j=1}^{d_{l-1}}\frac{\sum_{k=1}^{d_{l}} v_{l,j,k}^{2}}{\frac{1}{d_{l}}\sum_{k=1}^{d_{l}}\mathbb{E}_{v | y, \lambda}^{2}[v_{l,j,k}] + \frac{1}{d_{l}}\sum_{k=1}^{d_{l}}\text{Var}_{v | y, \lambda}[v_{l,j,k}]}.
\end{split}\end{equation}  Given the Gaussian prior assumption, what results is a sparsity-inducing $L_2$ penalty whose strength is inversely proportional to two factors: the squared mean and variance of the weight under the posterior.  The posterior mean can be thought of as \textit{signal}, the strength of the weight, and the variance can be thought of as \textit{robustness}, the scale invariance of the weight.

To further analyze the properties of (\ref{r_term_marg}), let us assume the current values of the weights are near their posterior means: $v_{l,j,k} \approx \mathbb{E}[v_{l,j,k}]$.  This assumptions simplifies (\ref{r_term_marg}) to \begin{equation}\label{r_term_marg_simplified}\begin{split}
\mathcal{R}_{GSM}(\mathbf{V}) &\approx \frac{1}{\sigma_{0}^{2}}\sum_{l=2}^{L}\sum_{j=1}^{d_{l-1}}\frac{d_{l}}{1 + \frac{\sum_{k=1}^{d_{l}}\text{Var}_{v | y, \lambda}[v_{l,j,k}]}{\sum_{k=1}^{d_{l}} v^{2}_{l,j,k}}}.
\end{split}\end{equation}  The fractional term in the denominator, $\sum_{k=1}^{d_{l}}\text{Var}_{v | y, \lambda}[v_{l,j,k}] / \sum_{k=1}^{d_{l}} v^{2}_{l,j,k}$, represents two alternative paths the weights can take to reduce the penalty during training.  The DNN must either send $\sum_{k=1}^{d_{l}} v^{2}_{l,j,k} \rightarrow 0$ or $\sum_{k=1}^{d_{l}}\text{Var}_{v | y, \lambda}[v_{l,j,k}] \rightarrow \infty$.  The former occurs when weights become sparse, and the latter occurs when weights are robust to rescaling (i.e. they do not have to be finely calibrated).  Hence, we observe a dual effect not seen in traditional sparsity penalties.  MN allows weights to grow without restraint just so long as they are invariant to rescaling.  If not, they are shrunk to zero.

Thinking back to how MN regularization is usually carried out in
practice (namely, by Monte Carlo sampling within the likelihood), we see that training in this way is essentially finding the invariant weights by brute force.  The only way the negative log likelihood can reliably be decreased is by pruning weights that cannot withstand being tested at random scales.  Dropout obscures this fact to some degree by being a discrete mixture over just two scales, zero and one.  The superior performance of continuous distributions, observed both in \cite{srivastava2014dropout, tomczak2013prediction} and further supported in our supplemental materials, may be due to searching over a richer, infinite scale space.

On a final note, the closed-form dropout penalties from equations (\ref{clt_dropout}) and (\ref{taylor_dropout}) can be recovered from $\mathcal{R}_{GSM}(\mathbf{V})$ by \textbf{1} assuming the Gaussian prior necessary for our analysis be diffuse and therefore negligible, and \textbf{2} the posterior mean is the same as the prior mean, which is necessary due to Wager et al. \cite{wager2013dropout} performing the Taylor expansion around the mean.  This removes the $\mathbb{E}^{2}[v]$ term from the denominator of $\mathcal{R}_{GSM}(\mathbf{V})$ (\ref{r_term_marg}).  Interestingly, this modification results in (\ref{r_term_marg}) becoming \begin{equation}\label{r_reg} \mathcal{R}_{GSMReg}(\mathbf{V}) = \frac{1}{\sigma_{0}^{2}}\sum_{j=1}^{d}\frac{v_{j}^{2}}{\text{Var}[v_{j}]}, \end{equation} which is the inverse of the term we isolated in Equation (\ref{r_term_marg_simplified}) as capturing the nature of MN regularization.  The resulting behavior is the same since we found the term in the denominator.  See the supplementary material for the details of the derivation.  Wager et al. interpreted their findings as an $L_{2}$ scaled by the inverse diagonal Fisher Information.  Yet, via the Cramer-Rao lower bound, their result could also be seen as an $L_{2}$ scaled by the \textit{asymptotic variance} of the weights.  A notion of variance, then, is just as integral to their frequentist derivation as it is our Bayesian one.

\section{Experiments: Weight Pruning}
We conducted a number of experiments to empirically investigate if our results present new directions for algorithmic improvements in training DNNs.
We implemented the EM algorithm derived in Section \ref{em_alg_sec} using Langevin Dynamics \cite{welling2011bayesian}, an efficient stochastic gradient technique for collecting posterior samples, to calculate the posterior moments needed for the M-Step.  We found that we could not outperform Monte Carlo MN regularization for any of the deep architectures with which we experimented (see supplemental materials).  We conjecture that the practical issue of computing the posterior moments was likely the bottleneck, which is to be expected given that developing efficient Bayesian learning algorithms for DNNs   is a challenging and  open problem in and of itself \cite{hernandez2015probabilistic, blundell2015weight}.  
%See the supplementary materials for a graph comparing the performance of training a 500-300 two-hidden layer network on MNIST with Monte Carlo MN (Bernoulli and Beta), an $L_{2}$ penalty, and EM.

However, we did find immediate and practical benefits in the context of model  compression \cite{ba2014deep, hinton2014distilling}.
Our conclusions about how MN regularizes DNNs conspicuously differ from the \textit{signal-to-noise ratio} (SNR) for weight pruning tasks, as used  by \cite{graves2011practical} and more recently by \cite{blundell2015weight}.  With this in mind we carried out a series of weight pruning experiments for the dual purpose of validating our analysis and providing a novel weight pruning rule (that turns out to be superior to the SNR).

The SNR heuristic is defined by the following inequality: $
\frac{|\mu_{l,j,k}|}{\sigma_{l,j,k}} < \tau$ where $|\mu_{l,j,k}|$ is the absolute value of the posterior mean of weight $v_{l,j,k}$, $\sigma_{l,j,k}$ is the posterior standard deviation of the same weight, and $\tau$ is some positive constant.  Pruning is carried out by setting to zero all weights for which the inequality holds (i.e. $|\mu|/\sigma$ is below some threshold $\tau$).  Blundell et al. \cite{blundell2015weight} ran experiments using the SNR and stated it ``is in fact related to test performance."

Now consider our alternative method.  Recall that the terms in the denominator of Equation (\ref{r_term_marg}) are $\mathbb{E}^{2}[v]+\text{Var}[v]$.  Our analysis shows that MN deems weights with large means \textit{and} large variances as being high quality, turning off the sparsity penalty applied to them.  This conclusion conflicts with the SNR since using $|\mu| / \sigma$  prunes weights with large variances first.  Thus we propose the following competing heuristic we call \textit{signal-plus-robustness} (SPR): \begin{equation}\label{spr} |\mu_{l,j,k}| + \sigma_{l,j,k} < \tau \end{equation} where the terms are defined the same as above.

\begin{figure}
\centering
\begin{subfigure}{.49\textwidth}
  \centering
  \includegraphics[width=.84\linewidth]{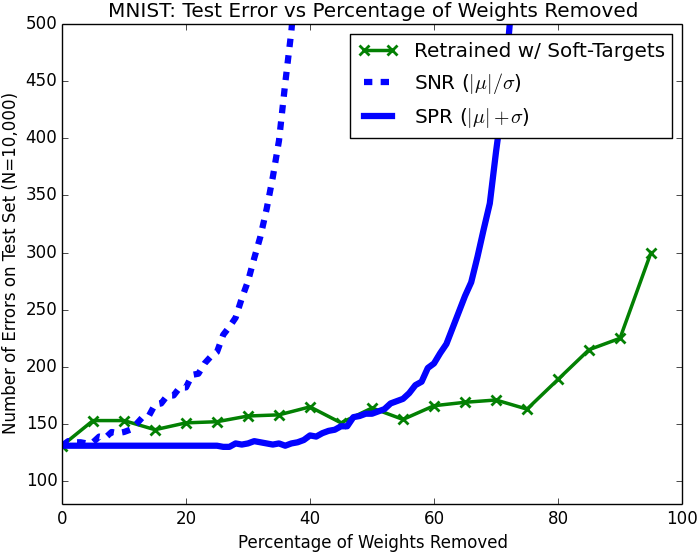}
  \caption{MNIST, 500-300 Hidden Units}
  \label{fig:sub1p}
\end{subfigure}
\begin{subfigure}{.49\textwidth}
  \centering
  \includegraphics[width=.84\linewidth]{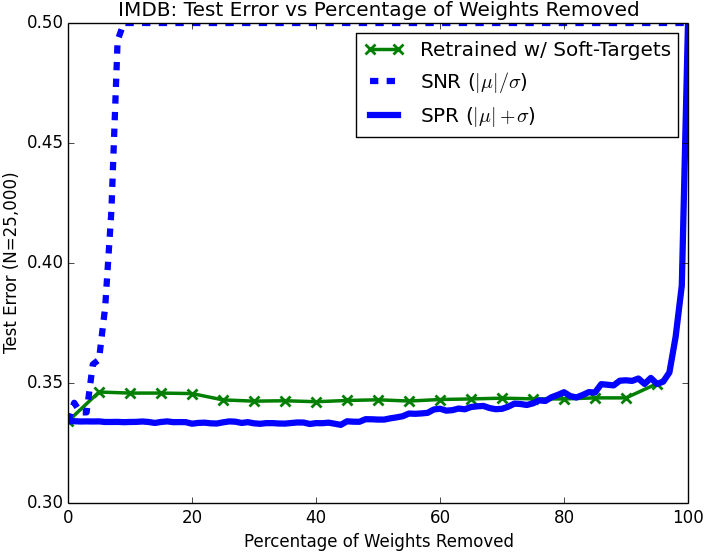}
  \caption{IMDB, 1000 Hidden Units}
  \label{fig:sub1p}
\end{subfigure}
\begin{subfigure}{.49\textwidth}
  \centering
  \includegraphics[width=.84\linewidth]{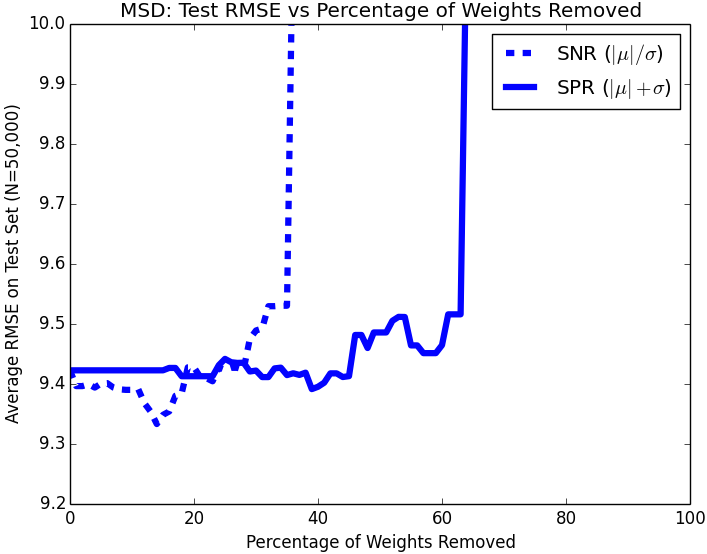}
  \caption{MSD, 120 Hidden Units}
  \label{fig:sub1p}
\end{subfigure}
\begin{subfigure}{.49\textwidth}
  \centering
  \includegraphics[width=.84\linewidth]{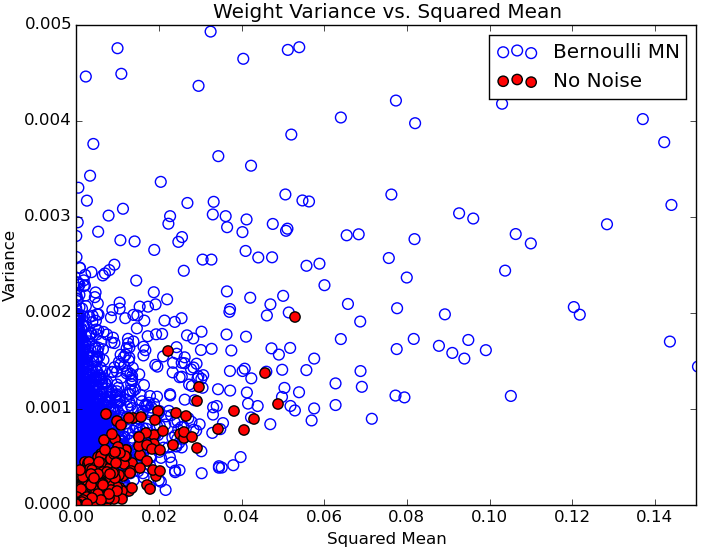}
  \caption{Posterior Weight Moments}
  \label{fig:sub0p}
\end{subfigure}
\caption{Experimental results: weight pruning task (a,b,c) and empirical moments (d).}
\label{experis}
\end{figure}
We experimentally compared both pruning rules on three datasets, each with very different characteristics.  The first is the well-known \textbf{MNIST} dataset ($d=784$, $N=50k/10k$), the second is the large \textbf{IMDB} movie review dataset for sentiment classification \cite{maas-EtAl:2011:ACL-HLT2011} ($d=5000$, $N=25k/25k$), and the third is a  prediction (regression) task using features preprocessed from the Million Song Dataset (\textbf{MSD}) \cite{Lichman:2013} ($d=90$, $N=460k/50k$).  We trained the networks with Bernoulli MN and when convergence was reached, switched to Langevin Dynamics (with no MN) to collected 10,000 samples from the posterior weight distribution of each network \cite{welling2011bayesian} ($\epsilon \sim N(0, lr/2$) where $lr$ is the learning rate).  A polynomial decay schedule was set by validation set performance.  

We ordered the weights of each network by SNR and SPR and then removed weights (i.e. set them to zero) in increasing order according to the two rules.  Plots showing test error (number of errors, error rate, mean RMSE) vs. percentage of weights removed can be seen in panels (a),(b) and (c) of Figure \ref{experis}.  For another source of comparison, we also show the performance of a network (completely) retrained on the soft-targets \cite{hinton2014distilling} produced by the full network\footnote{No soft-target results are shown for (c), the MSD year prediction task, as we found training with soft-targets does not have the same benefits for regression it does for classification}.  To make comparison fair, the retrained networks had the same depth as the one on which pruning was done, splitting the parameters equally between the layers.  

We see that our rule, SPR ($|\mu| + \sigma$), is clearly superior to SNR ($|\mu|/ \sigma$).  We were able to remove at least $20\%$ more of the weights in each case before seeing a catastrophic increase in test error.  The most drastic difference is seen for the IMDB dataset  in (b), which we believe is due to the sparsity of the features (word counts), exaggerating SNR's preference for overdetermined weights.  Our method, SPR, even outperformed retraining with soft-targets until at least a $50\%$ reduction in parameters was reached.  Finally, further empirical support of our findings, a scatter plot showing the first two moments of each weight for two networks--one trained with Bernoulli MN and the other without MN--can be see in panel (d) of Figure \ref{experis}.  We produce the figure to show that although our closed-form penalty technically doesn't hold for discrete noise distributions (due to the need to compute the gradient), the analysis (sparsity vs scale robustness) most likely extends to discrete mixtures.    

\section{Conclusions}
This paper improves our understanding of how multiplicative noise regularizes the weights of deep neural networks.  We show that multiplicative noise can be interpreted as a Gaussian scale mixture (under mild assumptions).  This perspective not only holds for neural networks regardless of their depth or activation function but allows us to isolate, in closed-form, the weight properties encouraged by multiplicative noise.  From this penalty we see that under multiplicative noise, the network's weights become either sparse or invariant to rescaling.  We demonstrated the utility of our findings by showing that a new weight pruning rule, naturally derived from our analysis, is significantly more effective than the previously proposed signal-to-noise ratio and is even competitive to retraining with soft-targets.

\bibliographystyle{plain}
\bibliography{references}

\appendix

\end{document}